%% file: fota.tex
\def\year{2022}\relax
\documentclass[letterpaper]{article} 
\usepackage{aaai22}  
\usepackage{times}  
\usepackage{helvet}  
\usepackage{courier}  
\usepackage[hyphens]{url}  
\usepackage{graphicx} 
\usepackage{natbib}  
\usepackage{caption} 
\DeclareCaptionStyle{ruled}{labelfont=normalfont,labelsep=colon,strut=off} 
\usepackage{algorithm}
\usepackage{algpseudocode}

\usepackage[dvipsnames]{xcolor}
\usepackage{subfiles}
\usepackage{amsmath}
\usepackage{amsthm}
\usepackage{amsfonts}
\usepackage{amssymb}
\usepackage{caption}
\usepackage{subcaption}
\usepackage{booktabs}
\usepackage{multirow}
\usepackage{makecell}
\usepackage{enumerate}
\usepackage{thmtools,thm-restate}
\include{lib/symbol}

\include{lib/consistency}

\newcommand{\name}{FOTA}
\newcommand{\src}{{\operatorname{s}}}
\newcommand{\tar}{{\operatorname{t}}}

\title{From One to All: Learning to Match Heterogeneous\\ and Partially Overlapped Graphs}
\author {
    Weijie Liu,\textsuperscript{\rm 1,2,3}
    Hui Qian,\textsuperscript{\rm 2,3,6}
    Chao Zhang,\textsuperscript{\rm 2,3}
    Jiahao Xie,\textsuperscript{\rm 2,3}
    Zebang Shen, \textsuperscript{\rm 4}
    Nenggan Zheng \textsuperscript{\rm 1,2,3,5\thanks{Corresponding author.}}
}
\affiliations {
    \textsuperscript{\rm 1} Qiushi Academy for Advanced Studies, Zhejiang University\\
    \textsuperscript{\rm 2} College of Computer Science and Technology, Zhejiang University\\
    \textsuperscript{\rm 3} Alibaba-Zhejiang University Joint Research Institute of Frontier Technologies\\
    \textsuperscript{\rm 4} University of Pennsylvania\\
    \textsuperscript{\rm 5} Zhejiang Lab\\
    \textsuperscript{\rm 6}State Key Lab of CAD\&CG, Zhejiang University\\
    \{westonhunter,qianhui,zczju,xiejh\}@zju.edu.cn, zebang@seas.upenn.edu, zng@cs.zju.edu.cn
}

\begin{document}
\maketitle
\begin{abstract}
	Recent years have witnessed a flurry of research activity in graph matching, which aims at finding the correspondence of nodes across two graphs and  lies at the heart of many artificial intelligence applications.
	However, matching heterogeneous graphs with partial overlap remains a challenging problem in \realworld{} applications.
	This paper proposes the first practical learning-to-match method to meet this challenge.
	The proposed unsupervised method adopts a novel partial OT paradigm to learn a transport plan and node embeddings simultaneously. 
	In a \emph{from-one-to-all} manner, the entire learning procedure is decomposed into a series of easy-to-solve \subprocedures{},	each of which only handles the alignment of a single type of nodes.
	A mechanism for searching the \emph{transport mass} is also proposed.
	Experimental results demonstrate that the proposed method outperforms state-of-the-art graph matching methods.
\end{abstract}

\section{Introduction}
\subfile{sections/introduction.tex}

\section{Preliminaries}\label{sec:preliminary}
\subfile{sections/preliminary.tex}

\section{Methodology}\label{sec:method}
\subfile{sections/method.tex}


\section{Experiments}\label{sec:experiment}
\subfile{sections/experiment.tex}

\section{Related Work}\label{sec:related}
\subfile{sections/related}

\section*{Conclusion}
In this paper, we propose the first practical method to match both heterogeneous graphs and partially overlapped graphs.
The learning procedure is decomposed into a series of \subprocedures{}, each of which matches one type of nodes by solving a partial optimal transport problem.
The nodes that are already matched serve as seeds.
Such a matching strategy is a hybrid of the \emph{seed-and-extend} strategy and the \emph{searching} strategy.
Empirical results demonstrate that our method outperforms state-of-the-art graph matching methods on both homogeneous and heterogeneous graphs.

\section*{Acknowledgments}
This work is supported by National Key Research and Development Program of China  under Grant 2020AAA0107400, Zhejiang Provincial Natural Science Foundation of China (Grant No: LZ18F020002, LR19F020005), Alibaba-Zhejiang University Joint Research Institute of Frontier Technologies, and National Natural Science Foundation of China (Grant No: 61672376, 61751209, 61472347, 61972347).

\bibliography{ref}

%
%
%
%
%
%
%
%
%
%
%
%

\end{document}

%% file: lib/symbol.tex
\newcommand{\AB}{\mathbf{A}}
\newcommand{\BB}{\mathbf{B}}
\newcommand{\CB}{\mathbf{C}}
\newcommand{\DB}{\mathbf{D}}

\newcommand{\KB}{\mathbf{K}}

\newcommand{\TB}{\mathbf{T}}

\newcommand{\WB}{\mathbf{W}}

\newcommand{\YB}{\mathbf{Y}}
\newcommand{\ZB}{\mathbf{Z}}

\newcommand{\zeroB}{\mathbf{0}}
\newcommand{\oneB}{\mathbf{1}}

\newcommand{\bB}{\mathbf{b}}

\newcommand{\pB}{\mathbf{p}}
\newcommand{\qB}{\mathbf{q}}

\newcommand{\xB}{\mathbf{x}}
\newcommand{\yB}{\mathbf{y}}
\newcommand{\zB}{\mathbf{z}}

\newcommand{\RBB}{\mathbb{R}}

\newcommand{\AM}{\mathcal{A}}

\newcommand{\DM}{\mathcal{D}}
\newcommand{\EM}{\mathcal{E}}

\newcommand{\GM}{\mathcal{G}}

\newcommand{\OM}{\mathcal{O}}
\newcommand{\PM}{\mathcal{P}}

\newcommand{\TM}{\mathcal{T}}
\newcommand{\VM}{\mathcal{V}}
\newcommand{\XM}{\mathcal{X}}
\newcommand{\YM}{\mathcal{Y}}
\newcommand{\Constraint}{\Omega}

\newcommand{\muB}{\mbox{\boldmath$\mu$\unboldmath}}

\newcommand{\zetaB}{\mbox{\boldmath$\zeta$\unboldmath}}

%% file: lib/consistency.tex
\newcommand{\recall}{\text{recall}}
\newcommand{\precision}{\text{precision}}
\newcommand{\Fone}{\text{F1}}

\newcommand{\dataset}{dataset}
\newcommand{\datasets}{datasets}
\newcommand{\groundtruth}{ground truth}

\newcommand{\realworld}{real-world}

\newcommand{\subgraphs}{sub-graphs}

\newcommand{\subproblem}{sub-problem}

\newcommand{\subprocedure}{sub-procedure}
\newcommand{\subprocedures}{sub-procedures}

\newcommand{\Dataset}{Dataset}
\newcommand{\Datasets}{Datasets}

%% file: sections/introduction.tex

Graph matching (network alignment), aiming to determine the correspondence of nodes across two related graphs, lies at the heart of a wide range of artificial intelligence applications, including network retrieval \cite{berretti2001efficient,ozer2002graph}, machine translation \cite{bahdanau2014neural,chen2020graph}, and visual tracking \cite{xiong2012structured,wang2017gracker}, to name a few. 
Two main algorithmic components, the \emph{node conservation} and the \emph{matching strategy}, constitute the graph matching method, where the former measures the similarity between pairs of nodes from different networks, and the later maximizes total node conservation over aligned nodes and the amount of conserved edges \cite{gu2018homogeneous}.

Mainstream methods of graph matching have followed two directions that correspond to different matching strategies.
{\bf (i)~}One is the \emph{seed-and-extend} strategy, which recursively matches nodes adjacent to the currently aligned subgraphs \cite{narayanan2009anonymizing,pedarsani2011privacy,yartseva2013performance,sun2015simultaneous}.
In general, incorrectly matched pairs could cause more mismatching in future steps, i.e., a cascade of errors \cite{kazemi2016network}.
{\bf (ii)~}The other is the \emph{searching} strategy, which explores the entire alignment space and returns the one with the highest score based on a specific objective.
In this paradigm, a few recent researches have shifted focus to settings in which  the best searching is rephrased in the framework of \emph{optimal transport} (OT) and the resulting algorithms have achieved the state-of-the-art performance \cite{xu2019gromov,titouan2019optimal,barbe2020graph}, compared with its traditional counterparts such as \emph{simulated annealing} \cite{mamano2017sana}, \emph{genetic algorithm} \cite{vijayan2015magna++}, or \emph{gradient-based optimization} \cite{konar2020graph}.

\begin{figure}[ht]
	\includegraphics[width=\linewidth]{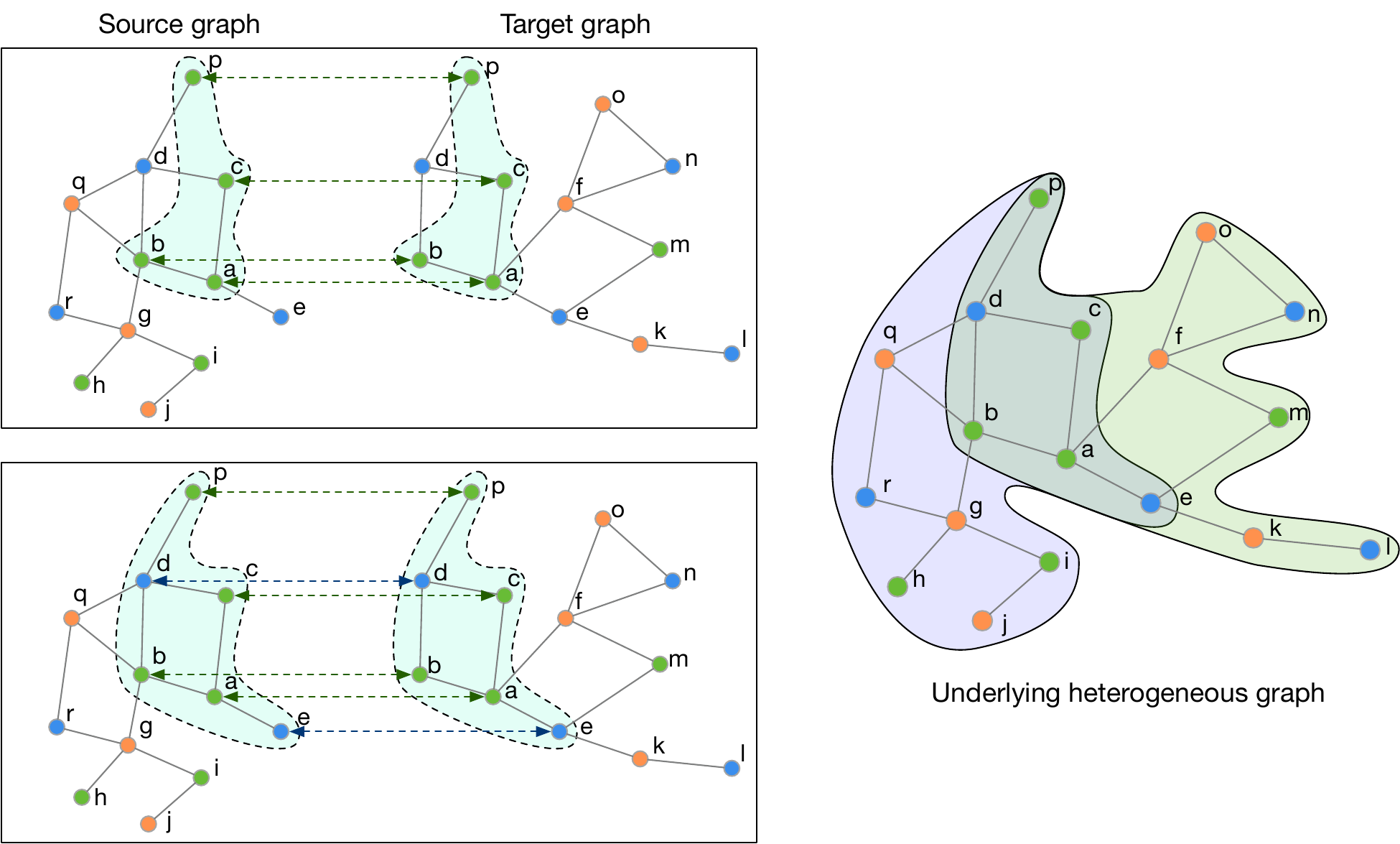}
	\caption{An illustration of the from-one-to-all matching strategy for both heterogeneous and partially overlapped graphs.
		The right sub-figure depicts an underlying heterogeneous graph.
		The color of each node indicates its type.
		The source graph and the target graph pictured in the left sub-figure are its two partially overlapped \subgraphs{}.
		For example, node r in the source graph does not exist in the target graph. First, type-green nodes are matched (top row of the left sub-figure). Then, type-blue nodes are matched (bottom row of the left sub-figure).}\label{fig:two_issues}
\end{figure}

Despite this recent activity,  the advances in methodology have been confined almost exclusively to the matching of \emph{homogeneous}  (containing only one type of nodes) or fully overlapped graphs,  and none of existing methods has demonstrated its effectiveness on real-world alignment tasks in which graphs are both \emph{heterogeneous}  and \emph{partially overlapped}.  
Such real-world tasks may include knowledge graph alignment \cite{li2018non}, matching of biological networks \cite{sharan2006modeling}, and disambiguating entities \cite{zhang2019shne}. 
We illustrate this situation in Figure \ref{fig:two_issues} and depict technical details of these two challenges in the following paragraphs.


\paragraph{Heterogeneity.}
Most existing methods consider graphs with only one type of nodes.
Directly applying these methods to heterogeneous graphs leads to inferior performance and type mismatch.
\citeauthor{gu2018homogeneous}~(\citeyear{gu2018homogeneous}) propose the first practical heterogeneous graph matching method based on manually designed features, called colored graphlet degree vectors (CGDV), and extend a large body of homogeneous algorithms \cite{sun2015simultaneous,vijayan2015magna++,mamano2017sana} into heterogeneous variants. The designed features, however, are expensive to obtain since the computational complexity for generating CGDV is practically $\OM(V^5)$, where $V$ is the number of nodes of the graph.
Apparently, as $V$ increases, their performances will deteriorate rapidly.
Besides, these methods cannot guarantee that there is no type mispairing. Thus, efficiently and effectively matching all types of nodes still remains a challenging problem.

\paragraph{Partial Overlapping.}
Partial overlapping is another challenging problem, which is generally tackled by adding \emph{dummy nodes} that act like wildcards to absorb unmatched nodes \cite{swoboda2017study,swoboda2019convex,sarlin2020superglue,rolinek2020deep}.
However, this kind of methods still suffer from two intractable issues.
First, they are usually supervised methods and require a large amount of ground truth node pairs.
Second, they cannot constrain the number of matching pairs even if the prior information about the degree of overlap is already known, which may lead to some extreme cases.
For example, zero matching would be identified to be an optimal result when dummy nodes are adopted.

To meet these challenges from real-world tasks,  we propose a novel method to match heterogeneous graphs that are possibly partially overlapped.
The proposed method adopts a novel partial OT scheme to learn a transport plan and node embeddings simultaneously. 
For the sake of computational efficiency, the entire learning procedure is decomposed into a series of easy-to-solve \subprocedures{}. Specifically,  each \subprocedure{} only handles a homogeneous but possibly partially overlapped alignment problem, with a set of node pairs which are already matched in the previous \subprocedures{} as seeds.  
Essentially, this method proceeds by first matching one type of nodes and gradually matching other types until all types are matched, that is, from one to all types, so we name it \name{}.
To boost robustness against noise and further reduce the cascade of errors, we also incorporate node embeddings which encode the global topology.
Our contributions are summarized as follows.
\begin{enumerate}[i.]
	\item \name{} is the first practical unsupervised method to match both heterogeneous and partially overlapped graphs,  so far as we know, by using a hybrid strategy that mixes seed-and-extend and searching strategies. Compared to the principled method in \cite{gu2018homogeneous}, ours achieves no type mismatch at a very low cost.
	The overall complexities for updating the transport plan and the embeddings are $\OM\big(N(T+k+d)V^2\big)$ and  $\OM(VBd)$ respectively\footnote{Here $N$, $T$, $k$, $d$ and $B$ denote the number of main iterations, the number of iterative projections, the rank of approximation to the proximity matrix, embedding dimension, and the batch size respectively. See Sec. \ref{sec:complexity} for detail.}.
	\item A mechanism for searching the transport mass is also proposed, which endows our method a possibility to control the number of matching pairs according to some prior information. 
	Thus the  partial OT technique leveraged by our method only needs to transport an enough fraction of the mass with a minimum transportation cost.  
\end{enumerate}
Extensive experimental results demonstrate that the proposed method outperforms the state-of-the-art
graph matching methods.
The rest of the paper is organized as follows.
In Sec. \ref{sec:preliminary}, a comprehensive review of background is given.
The methodology of \name{} is presented in Sec. \ref{sec:method}.
Empirical results are demonstrated  in both Sec. \ref{sec:experiment}.
We finally present related work in Sec. \ref{sec:related}.


\paragraph{Notation.}
We use bold lowercase symbols, bold uppercase letters, uppercase calligraphic fonts, and Greek letters to denote vectors, matrices, spaces (sets), and measures, respectively.
$\oneB^d\in\RBB^d$ is an all-ones vector.
The cardinality of set $\AM$ is denoted by $|\AM|$.
$\AB[i,:]$ and $\AB[:,j]$ are the $i$-th row and the $j$-th column of matrix $\AB$ respectively.

%% file: sections/preliminary.tex
\subsection{Graph Matching}
A graph is denoted as $\GM=(\VM,\EM,\phi,\TM)$, where $\VM$ is the set of nodes, $\EM$ is the set of edges, $\TM$ is the set of node types, and the type mapping function $\phi:\VM\to\TM$ assigns each node a type.
When graph $\GM$ contains multiple types of nodes, i.e., $|\TM|\ge 2$, $\GM$ is a heterogeneous graph;
otherwise, it is a homogeneous graph.
Assigning each node an index $i\in\{1,\dots,|\VM|\}$, the edge set $\EM$ can also be written as an adjacency matrix $\WB\in\RBB^{|\VM|\times|\VM|}$ with entry $W_{ij}=1$ if and only if there is an edge connecting nodes $i$ and $j$.

Mathematically, graph matching finds a matching matrix $\TB=[T_{ii'}]$ between the source graph $\GM^\src=(\VM^\src,\EM^\src,\phi^\src,\TM)$ and the target graph $\GM^\tar=(\VM^\tar,\EM^\tar,\phi^\tar,\TM)$,
where $T_{ii'}=1$ if node $i\in\VM^\src$ is matched to node $i'\in\VM^\tar$, and $T_{ii'}=0$ otherwise.
Assuming $\GM^\src$ has fewer nodes than $\GM^\tar$,
the matching matrix is generally identified by minimizing the following loss function \cite{caetano2009learning}
\begin{equation}\label{eq:formulation}
	\min_{\TB\in\PM^f}\bigg[\sum_{i,j=1}^{|\VM^\src|}\sum_{i',j'=1}^{|\VM^\tar|}d_{ii'jj'}T_{ii'}T_{jj'}+\sum_{i=1}^{\VM^\src|}\sum_{i'=1}^{|\VM^\tar|}k_{ii'}T_{ii'}\bigg],
\end{equation}
where $k_{ii'}$ is the cost for the \emph{unary matching} $i\rightarrow i'$, $d_{ii'jj'}$ is the cost for the \emph{pairwise matching} $(i,j)\rightarrow (i',j')$,
and the feasible domain is
\begin{equation*}
	\resizebox{.91\linewidth}{!}{$
		\displaystyle
		\PM^f=\Big\{\TB\in\{0,1\}^{|\VM^\src|\times |\VM^\tar|}\Big|\TB\oneB^{|\VM^\tar|}=\oneB^{|\VM^\src|},\TB^\top\oneB^{|\VM^\src|}\le\oneB^{|\VM^\tar|}\Big\}.
	$}
\end{equation*}
Such formulation matches every node in $\GM^\src$ to exact one node in $\GM^\tar$ and hence is referred to as \emph{full matching} \cite{kazemi2015can}.
For \emph{partial matching}, generally $\PM^f$ is replaced with the relaxed feasible domain
\begin{equation*}
	\resizebox{.91\linewidth}{!}{$
		\displaystyle
			\PM^p=\Big\{\TB\in\{0,1\}^{|\VM^\src|\times |\VM^\tar|}\Big|\TB\oneB^{|\VM^\tar|}\le\oneB^{|\VM^\src|},\TB^\top\oneB^{|\VM^\src|}\le\oneB^{|\VM^\tar|}\Big\},
		$}
\end{equation*}
see e.g. \cite{kazemi2015can,sarlin2020superglue,wang2020zero}.
Such formulation, however, may lead to matching no pairs in extreme cases.
For example, when $d_{ii'jj'}>0$ and $k_{ii'}>0$ for all $i,i'j,j'$, matching no nodes ($\TB=\zeroB$) is the optimal solution.
Matching two heterogeneous graphs further requires $\phi^\src(i)=\phi^\tar(i')$ if node $i$ is matched to $i'$.

\subsection{Optimal Transport}

Optimal Transport (OT) addresses the problem of transporting one measure toward another measure with the minimum cost \cite{villani2008optimal}.
The induced cost defines a distance between the two measures.
A discrete measure $\alpha$ can be denoted by $\alpha=\sum_{i=1}^{m}p_i\delta_{\xB_i}$ where $\delta_{\xB}$ is the Dirac at position $\xB$, i.e., a unit of mass infinitely concentrated at $\xB$.
With slight abuse of notation, we also use $\pB=[p_i]$ to refer to $\alpha$.

\paragraph{Wasserstein distance.}
The $p$-Wasserstein distance \cite{villani2008optimal,cuturi2013sinkhorn} between discrete measures $\pB$ and $\qB$ is defined as
\begin{equation*}\textstyle
	W_p^p(\pB,\qB)=\min_{\TB\in\Pi(\pB,\qB)}\sum_{i=1}^{m}\sum_{i'=1}^{n} K^p_{ii'}T_{ii'},
\end{equation*}
where $K_{ii'}$ is the $\ell_p$ distance between $\xB_i$ and $\yB_{i'}$ and the feasible domain of \emph{transport plan} $\TB=[T_{ii'}]$ is given by the set $\Pi(\pB,\qB)=\{\TB\in\RBB_+^{m\times n}|\TB\oneB^n=\pB,\TB^\top\oneB^m=\qB\}$.
The Wasserstein distance requires the supports of the two measures to be in the same space.

\paragraph{Gromov-Wasserstein distance.}~
Gromov-Wasserstein (GW) distance extends Wasserstein distance to compare measures supported in different spaces \cite{memoli2011gromov}.
Let $\XM$ and $\YM$ be two sample spaces.
Endowing the spaces $\XM$ and $\YM$ with metrics (distances) $d_\XM$ and $d_\YM$,
the GW distance is defined as
\begin{equation*}
	GW_p^p(\pB,\qB)=\min_{\TB\in\Pi(\pB,\qB)}\sum_{i,j=1}^{m}\sum_{i',j'=1}^{n}D_{ii'jj'}^pT_{ii'}T_{jj'},
\end{equation*}
where $D_{ii'jj'}=|d_\XM(\xB_i,\xB_j)-d_\YM(\yB_{i'},\yB_{j'})|$  with $\xB_1$, $\xB_2$, $\ldots$, $\xB_m\in\XM$ and $\yB_1$, $\yB_2$, $\ldots$, $\yB_n\in\YM$.

\paragraph{OT-based graph matching.}
By associating each graph with a discrete probability measure, OT  can be applied to graph matching.
\citeauthor{xu2019gromov}~(\citeyear{xu2019gromov}) propose a GW learning framework called GWL for graph matching.
They correspond two graphs $\GM^\src$ and $\GM^\tar$ to discrete probability measures $\muB^\src=[\mu^\src_i]$ and $\muB^\tar=[\mu^\tar_i]$ respectively,
where 
\[
\mu^z_i=\frac{\sum_{j=1}^{|\VM^z|} W_{ij}^z}{\sum_{i=1}^{|\VM^z|}\sum_{j=1}^{|\VM^z|}W_{ij}^z}, 
\]
for $z=\src,\tar$.
By replacing the strict distances with dissimilarity functions, GWL relaxes the GW distance and the Wasserstein distance to the GW discrepancy and the Wasserstein discrepancy separately.
Such relaxation allows GWL to incorporate node embeddings to parameterize the discrepancies and improve the robustness to the noise of edges.
GWL uses the learned transport plan to indicate the node correspondence, i.e., $i'\in\VM^\tar$ that receives the most mass from $i\in\VM^\src$ is the estimated counterpart of $i$.


Both the Wasserstein distance and the GW distance require the two marginals $\pB$ and $\qB$ to have the same total mass, that is, $\|\pB\|_1 = \|\qB\|_1$, thus all the mass has to be transported.
By contrast, the partial OT problem focuses on transporting only a fraction $0\le b\le \min\{\|\pB\|_1, \|\qB\|_1\}$ of the mass with the minimum transportation cost \cite{figalli2010optimal,caffarelli2010free,chapel2020partial}, that is, the set of admissible couplings is given by
\begin{equation*}
	\Pi^b(\pB,\qB)=\Big\{\TB\in\RBB_+^{m\times n}\Big|\TB\oneB\le\pB,\TB^\top\oneB\le\qB,\oneB^\top\TB\oneB=b\Big\}.
\end{equation*}
By adding a dummy node into the target graph, the mass of the nodes which have no counterparts in the target graph can be considered as being transported to the dummy node.

%% file: sections/method.tex
We first introduce the underlying model in \realworld{} scenarios for matching graphs that are heterogeneous and partially overlapped.
Next, we derive a practical \emph{from-one-to-all} model by decomposing the learning procedure into a series of interrelated \subprocedures{}.
A recursive line search mechanism is then proposed to search for the transport mass in order to conduct partial matching.
Finally, we analyze the overall complexity of the proposed method.

\subsection{Proposed Model}\label{sec:overall_model}
Our model is a learning-to-match model which estimates the transport plan and the node embeddings simultaneously.
The optimization formulation can be phrased as follows:
\begin{equation}\label{eq:objective}
	\begin{aligned}
		\min_{\ZB^\src,\ZB^\tar}&\min_{\TB\in\Constraint(\bB,\GM^\src,\GM^\tar)}\underbrace{\sum_{i,j=1}^{|\VM^\src|}\sum_{i',j'=1}^{|\VM^\tar|}(C_{ij}^\src-C_{i'j'}^\tar)^2T_{ii'}T_{jj'}}_{\text{partial GW discrepancy}}\\
		&+\alpha\underbrace{\sum_{i=1}^{|\VM^\src|}\sum_{i'=1}^{|\VM^\tar|}-\kappa(\zB_i^\src,\zB_{i'}^\tar)T_{ii'}}_{\text{partial Wasserstein discrepancy}}+\beta \underbrace{\big(R(\ZB^\src)+R(\ZB^\tar)\big)}_{\text{regularizers}},
	\end{aligned}
\end{equation}
where $\ZB^\src=[\zB_i^\src]$ and $\ZB^\tar=[\zB_{i'}^\tar]$ are node embeddings of $\GM^\src$ and $\GM^\tar$ respectively, $\bB=[b_r]$ specifies the transport mass for each type.
The feasible domain $\Constraint(\bB, \GM^\src, \GM^\tar)$ is defined as
\begin{equation*}
	\begin{aligned}
		\Constraint(\bB, \GM^\src, \GM^\tar&)=\Big\{\TB\in\RBB_+^{|\VM^\src|\times|\VM^\tar|}\Big|\TB\oneB\le\muB^\src,\TB^\top\oneB\le\muB^\tar,\\
		&T_{ii'}=0\text{ if }\phi^\src(i)\neq\phi^\tar(i')\forall (i,i')\in\VM^\src \times\VM^\tar,\\
		&\textstyle\sum_{i:\phi^\src(i)=r}\sum_{i'=1}^{|\VM^\tar|}T_{ii'}=b_r~\forall r\in\TM\Big\},
	\end{aligned}
\end{equation*}
where the third constraint guarantees that nodes of different types are not matched and the fourth indicates that the transport mass for type $r$ is $b_r$.
$\alpha$ and $\beta$ are constant scalars and the three components of the objective are described as follows.

\paragraph{Partial GW Discrepancy.}
The partial GW discrepancy is adopted to measure the sum of the pairwise matching costs.
In heterogeneous graphs, nodes may not be adjacent to nodes of its own type.
To take into account the in-type topological structure, we consider multi-hop connections.
Specifically, we calculate a matrix that characterizes the $K$\textsuperscript{th}-order proximity as
\begin{equation}\label{eq:proximity matrix}\textstyle
	\BB^z=\sum_{l=1}^{K}\big((\DB^z)^{-1}\WB^z\big)^l+\big(\WB^z(\DB^z)^{-1}\big)^l,
\end{equation}
where $z=\src,\tar$ and $\DB^z=[D^z_{ij}]$ is the diagonal degree matrix with the $i$\textsuperscript{th} entry given by $D^z_{ii}=\sum_{j=1}^{|\VM^z|}W^z_{ij}$.
The $(i,j)$\textsuperscript{th} entry of matrix $\big((\DB^z)^{-1}\WB^z\big)^l$ is the probability that, at the $l$\textsuperscript{th} step, $j$ is visited by a random walk starting from $i$ \cite{qiu2018network}.
Similar explanation applies to the $(i,j)$\textsuperscript{th} entry of $\big(\WB^z(\DB^z)^{-1}\big)^l$.
We use \emph{proximity matrix} $\CB^z=[C_{ij}^z]$ to model the proximity between nodes $i$ and $j$ where
\begin{equation}\label{eq:constant_proximity}
	C_{ij}^z=\begin{cases}
		B_{ij}^z,&\text{ if }i\neq j,\\
		\bar{C},&\text{ otherwise,}
	\end{cases}
\end{equation}
where $z=\src,\tar$ and $\bar{C}$ is a large enough constant.

\paragraph{Partial Wasserstein Discrepancy.}
The partial Wasserstein discrepancy is the sum of the unary matching costs based on the learned node embeddings.
Herein, $\kappa(\zB_i^\src,\zB_{i'}^\tar)=(\zB_i^{\src\top}\zB_{i'}^\tar)/(\|\zB_i^\src\|\|\zB_{i'}^\tar\|)$ is the normalized inner-product between node embeddings $\zB_i^\src$ and $\zB_{i'}^\tar$ and measures the similarity between nodes $i$ and $i'$.
By incorporating node embeddings, we improve the robustness to the structure noise.
Note that, for attributed graphs in which each node is associated to a feature vector, heterogeneous graph neural networks (e.g., \cite{zhang2019heterogeneous}) can be used to incorporate attributes information and parameterize the node embeddings.
For simplicity, this paper only considers unattributed graphs.

\paragraph{Regularizers.}
$R(\ZB^z)$ for $z=\src$ or $\tar$, regularizes the embedding by penalizing the distance between the inner-product of node embeddings and the proximity, and is defined as
\begin{equation*}\textstyle
	R(\ZB^z)=\sum_{i=1}^{|\VM^z|}\sum_{j:\phi^z(i)=\phi^z(j),j\neq i}\big(C_{ij}^z-\zB_i^{z\top}\zB_j^z\big)^2.
\end{equation*}
The regularizers guarantee that node embeddings capture the global topological structure via preserving the total high-order proximity \cite{tang2015line,cao2015grarep}.

With the composition of the above three components, an appropriate algorithm can learn embedding vectors and node correspondence in a simultaneous manner.
This joint learning formula has been shown to substantially improve the robustness to the noise \cite{xu2019gromov,karakasis2021joint}.

\subsection{Model Decomposition: From One to All}\label{sec:model_decomp}
We decompose the learning procedure into a series of easy-to-solve \subprocedures{} due to the following facts.
{\bf (i)~}The complicated constraint $\Constraint(\bB, \GM^\src, \GM^\tar)$ makes the model corresponding to (\ref{eq:objective}) difficult to train.
{\bf (ii)~}In several scenarios such as user alignment \cite{wang2019user}, the matching of only a few primary types are of interest.

Before delving into the decomposition of the learning procedure, we first introduce some additional notations.
Without loss of generality, we assume the node types are represented by integers, i.e., type $r\in\{1,2,\dots,|\TM|\}$.
$\zetaB^{z,r}=\Big[\zeta^{z,r}_i\Big]$ is the mask vector for type $r$  and is defined as
\begin{equation*}
	\zeta^{z,r}_i=\begin{cases}
		1,&\text{ if }\phi^z(i)=r,\\
		0,&\text{ otherwise.}
	\end{cases}
\end{equation*}
Based on the mask vectors, $\TB$ can be written as the sum of $|\TM|$ matrices $\TB=\sum_{r=1}^{|\TM|}\TB^{r}$ where $\TB^{r}=\TB\odot\Big(\zetaB^{\src,r}\zetaB^{\tar,r\top}\Big)$ is the matching for type-$r$ nodes and $\odot$ is the element-wise multiplication.
Similarly, $\muB^\src$ and $\muB^\tar$ can be written as the sum of \emph{type measures}, i.e., $\muB^z=\sum_{r=1}^{|\TM|}\muB^{z,r}$ where the type-$r$ measure is $\muB^{z,r}=\muB^z\odot\zetaB^{z,r}$.
We further denote the set of type-$r$ nodes and the corresponding embeddings by $\VM^{z,r}=\{i|\phi^z(i)=r\}$ and $\ZB^{z,r}=\{\zB_{i}^z|\phi^z(i)=r\}$ respectively.

We now decompose the learning procedure for (\ref{eq:objective}) into $|\TM|$ \subprocedures{}, which is formally stated in the following proposition.

\begin{restatable}{proposition}{primeproposition}
	\label{prop:model_decomposition}
		The learning procedure (\ref{eq:objective}) can be decomposed into $|\TM|$ \subprocedures{}.
		The $r$\textsuperscript{th} \subprocedure{} corresponds to the learning of type-$r$ matching $\TB^{r}$ and the embeddings $\ZB^{\src,r}$ and $\ZB^{\tar,r}$.
		The optimization problem is $\min_{\ZB^{\src,r},\ZB^{\tar,r}}\min_{\TB^{r}}f^{r}(\TB^{r},\ZB^{\src,r},\ZB^{\tar,r})$ and the objective is given by
		\begin{equation}\label{eq:subproblem_objective}
			\begin{aligned}
				&f^{r}(\TB^{r},\ZB^{\src,r},\ZB^{\tar,r})\\
				=&\sum_{i,j=1}^{|\VM^\src|}\sum_{i',j'=1}^{|\VM^\tar|}(C_{ij}^\src-C_{i'j'}^\tar)^2T_{ii'}^{r}\Big(T_{jj'}^{r}+2\sum_{a=1}^{r-1}\hat{T}_{jj'}^{a}\Big)\\
				+&\alpha\sum_{i\in\VM^{\src,r}}\sum_{i'\in\VM^{\tar,r}}-\kappa(\zB_i^\src,\zB_{i'}^\tar)T_{ii'}^{r}+\beta \big(R^{r}(\ZB^{\src,r})+R^{r}(\ZB^{\tar,r})\big),
			\end{aligned}
		\end{equation}
		where $\TB^{r}\in\Pi^{b_r}\big(\muB^{\src,r},\muB^{\tar,r}\big)$, $\hat{\TB}^{a}=[\hat{T}_{jj'}^{a}]$ is the obtained matching for type-$a$ nodes, and the regularizers are $R^{r}(\ZB^{z,r})=\sum_{i,j\in\VM^{z,r},i\neq j}\big(C_{ij}^z-\zB_i^{z\top}\zB_j^z\big)^2, z=\src,\tar$.
\end{restatable}

The proof is deferred to the long version due to the limit of space.
Intuitively, to match each type, the nodes already matched serve as seeds.
Node embeddings that encode global topology are incorporated to enhance robustness to edge noise and reduce the cascade of errors.

$\Pi^{b_r}\big(\muB^{\src,r},\muB^{\tar,r}\big)$ is a relatively simpler feasible domain.
Hence, the \subproblem{} can be practically solved via alternating optimization, that is, alternatingly updating the transport plan and node embeddings, which is detailed as follows.

\paragraph{Updating the transport plan.}
Given current node embeddings $\ZB^{\src,r,(m)}$ and $\ZB^{\tar,r,(m)}$, we solve the following problem,
\begin{equation}\label{eq:pot}\textstyle
	\min_{\TB^{r}\in\Pi^{b_r}\big(\muB^{\src,r},\muB^{\tar,r}\big)}f^{r}\big(\TB^{r},\ZB^{\src,r,(m)},\ZB^{\tar,r,(m)}\big).
\end{equation}
Minimization (\ref{eq:pot}) is a partial OT problem and can be effectively solved using off-the-shelf OT algorithms.

\paragraph{Updating the embeddings.}
Given the calculated transport plan $\hat\TB^{r,(m)}$, we update the embeddings.
The \subproblem{} for updating embeddings is
\begin{equation}\label{eq:embedding subproblem}
	\begin{aligned}
		\min_{\ZB^{\src,r},\ZB^{\tar,r}}&\textstyle\alpha\sum_{i\in\VM^{\src,r}}\sum_{i'\in\VM^{\tar,r}}-\kappa(\zB_i^\src,\zB_{i'}^\tar)T_{ii'}^{r,(m)}\\\textstyle
		+&\beta \big(R^{r}(\ZB^{\src,r})+R^{r}(\ZB^{\tar,r})\big),
	\end{aligned}
\end{equation}
which can be solved by SGD or its variants.

We summarize \name{} in Algorithm \ref{alg:fota} and make some important remarks as follows.
\begin{enumerate}
	\item {\bf Extended transport plan.} The final learned transport plan $\hat{\TB}$ is used to construct an extended transport plan $\tilde{\TB}=[\hat{\TB},\muB^\src-\hat{\TB}\oneB]\in\RBB^{|\VM^\src|\times(|\VM^\tar|+1)}$.
	Nodes matched to the dummy node are considered to have no counterpart in the target graph.
	By choosing the largest $\tilde{T}_{ij}$ for each i, we find the correspondence.
	\item {\bf Refinement.} The current algorithm uses $|\TM|$ iterations to obtain the matching of each type, which is analogous to that of a single round of block coordinate descent.
	Since approximation errors exist in each iteration, empirically, we can repeat Line \ref{algline:iteration begin} to \ref{algline:iteration end} in Alg \ref{alg:fota} to refine the matching.
\end{enumerate}


\begin{algorithm}[ht]
	\caption{\name{}\label{alg:fota}}
	\begin{algorithmic}[1]
		\State {\textbf{Input:}} $M$ rounds, $K$, $\alpha$, $\beta$, graphs $\GM^\src$ and $\GM^\tar$.
		\State {\textbf{Output:}} Correspondence set $\DM$.
		\State Calculate $\CB^\src$ and $\CB^\tar$ as Eq. (\ref{eq:constant_proximity}).
		\State Obtain transport mass for each type $\bB$ via Algorithm \ref{alg:linesearch}.
		\State Initialize $\ZB^\src$ and $\ZB^\tar$ via SVD for $\CB^\src$ and $\CB^\tar$ respectively.
		
		\For{$r=1,\dots,|\TM|$}\label{algline:iteration begin}
		\For{$m=1,\dots,M$}
		\State Update transport plan by solving (\ref{eq:pot}).
		\State Update node embeddings by solving (\ref{eq:embedding subproblem}).
		\EndFor
		\State $\hat\TB^{r}=\hat\TB^{r,(M)}$.
		\EndFor\label{algline:iteration end}
		\State $\hat{\TB}=\sum_{r=1}^{|\TM|}\hat\TB^{r}$.
		\State $\tilde{\TB}=[\hat{\TB},\muB^\src-\hat{\TB}\oneB]$.
		\State Initialize correspondence set $\DM=\emptyset$.
		\For{$i\in\{1,\dots,|\VM^\src|\}$}
		\State $i'=\arg\max_{i'} \tilde{T}_{ii'}$.
		\If{$i'\neq(|\VM^\tar|+1)$}
		\State $\DM=\DM\cup\{(i,i')\}$.
		\EndIf
		\EndFor
	\end{algorithmic}
\end{algorithm}

\begin{algorithm}[ht]
	\caption{LineSearch\label{alg:linesearch}}
	\begin{algorithmic}[1]
		\State {\textbf{Input:}} type $r$, line search resolution $\delta$, number $n$, search range $(b_{\min}, b_{\max}]$, threshold $\gamma$.
		\State {\bf Output:} $\hat{b}_r$.
		\For{$i=1,\dots,n$}
		\State $b_i=b_{\min}+\frac{i}{n}(b_{\max}-b_{\min})$.
		\State Obtain the approximation $\hat{A}(b_i)$ for $A(b_i)$.
		\If{$\hat{A}(b_i)\ge\frac{1+\gamma}{i}\sum_{j=1}^{i}\hat{A}(b_j)$}
		\If{$b_{\max}-b_{\min}>\delta$}
		\State LineSearch($r$,$\delta$, $n$, $(b_{i-1},b_i]$, $\gamma$).
		\Else
		\State $\hat{b}_r=b_{\min}$.
		\State \Return
		\EndIf
		\EndIf
		\EndFor
	\end{algorithmic}
\end{algorithm}

\begin{table}[ht]
	\centering
	\small
	\begin{tabular}{ccccccccc}
		\toprule
		\multicolumn{2}{c}{\bf{\Datasets{}}}     & $\rho$=1.0 & $\rho$=0.8  & $\rho$=0.6 & \bf{Description}                            \\
		\midrule
		\multirow{4}{*}{\makecell[c]{\bf{Arenas} \\ \bf{Email}}} & $|\VM^\src|$ &  1133            &   986       &  844          & \multirow{4}{*}{\makecell[c]{email \\ network}} \\
		& $|\EM^\src|$ & 10902           &    8694   &    6760        &                                        \\
		& $|\VM^\tar|$ &    1133       &   992       &      868    &                                        \\
		& $|\EM^\tar|$ &    10902        &  9022        &      7346      &                                  \\
		\midrule
		\multirow{4}{*}{\makecell[c]{\bf{PPI} \\ \bf{Yeast}}} & $|\VM^\src|$ &  1004           & 835         &    628        & \multirow{4}{*}{\makecell[c]{protein \\ interaction}} \\
		& $|\EM^\src|$ 																							&  16646     &   13406  &   8474      &                                        \\
		& $|\VM^\tar|$ 																							&   1004       &  844            &   661          &                                        \\
		& $|\EM^\tar|$ 																							&   16646     & 13270     &    10396&                                  \\
		\midrule
		\multirow{4}{*}{\makecell[c]{\bf{Arxiv}}} & $|\VM^\src|$ & 18772         &  15667       & 13757           & \multirow{4}{*}{\makecell[c]{coauthor \\ network}} \\
		& $|\EM^\src|$ 																		&  396160     &   318846    &  261206      &                                        \\
		& $|\VM^\tar|$ 																		&   18772       &   15933         &   14073          &                                        \\
		& $|\EM^\tar|$ 																		&    396160    &    327403	  &   267646     &                                  \\
		\bottomrule     
	\end{tabular}
	\caption{Statistics of homogeneous graphs used in our experiments.}
	\label{tab:homo datasets}
\end{table}

\subsection{Transport Mass Search}\label{sec:linesearch}
Intuitively, the partial GW discrepancy is very small if the chosen $b_r$ is less than the underlying transport mass.
It soars when $b_r$ begins to exceed the underlying transport mass, since matching more non-overlapped nodes with disparate topology incurs larger penalty.
Therefore, for each type $r$, one can conduct a line search over $\big(0,\min\{\|\muB^{\src,r}\|_1,\|\muB^{\tar,r}\|_1\}\big]$ and find the turning point $\hat b_r$ of the partial GW discrepancy
\begin{equation}\label{eq:pure structure GW}
	A(b_r):=\min_{\TB\in\Pi^{b_r}}\sum_{i,j,i',j'}(C_{ij}^\src-C_{i'j'}^\tar)^2T_{ii'}T_{jj'}.
\end{equation}

Specifically, we adopt a recursive strategy.
We search the range $\big(b_{\min},b_{\max}\big]$ by evaluating $A(b_i)$ where $b_i=b_{\min}+\frac{i}{n}(b_{\max}-b_{\min})$ and $n$ is a preset number of samples.
When $A(b_i)>\frac{1+\gamma}{i}\sum_{j=1}^{i}A(b_j)$ where $\gamma$ is a given threshold, the optimal $\hat{b}_r$ is believed to fall into the range $\big(b_{i-1}, b_i\big]$.
Then we set $b_{\min}=b_{i-1}$ and $b_{\max}=b_i$, and repeat the above procedure until $b_{\max}-b_{\min}<\delta$ where $\delta$ is a small number.
In the beginning of the recursive line search, $b_{\min}=0$ and $b_{\max}=\min\{\|\muB^{\src,r}\|_1,\|\muB^{\tar,r}\|_1\}\big]$.
We summarize this line search strategy in Algorithm \ref{alg:linesearch}.


\begin{table*}[ht]
	\centering
	\small
	\begin{tabular}{cc ccc}
		\toprule
		\multicolumn{2}{c}{\textbf{Datasets}}        & $\rho$=1.0                                      & $\rho$=0.8 & $\rho$=0.6      \\
		\midrule
		\multirow{4}{*}{\bf{Movies}} & $|\VM^\src|$ & 0:348, 1:389, 2:257, 3:6 &   0:300, 1:306, 2:202, 3:3      &    0:237, 1:234, 2:159, 3:3      \\
		& $|\EM^\src|$ & 4618                                         &  3476       &      2612                     \\
		& $|\VM^\tar|$ & 0:348, 1:389, 2:257, 3:6 &  0:307, 1:329, 2:212, 3:3       &  0:261, 1:230, 2:150, 3:4                       \\
		& $|\EM^\tar|$ & 4618                                         & 3986        &      3080                    \\
		\midrule
		\multirow{4}{*}{\bf{PubMed}} & $|\VM^\src|$ & 0:1059, 1:1096, 2:1176, 3:669 &   0:908, 1:944, 2:1028, 3:572      &    0:808, 1:801, 2:908, 3:520     \\
		& $|\EM^\src|$ & 18982                                    &  15895     &      13527                   \\
		& $|\VM^\tar|$ & 0:1059, 1:1096, 2:1176, 3:669 &  0:925, 1:966, 2:1038, 3:591       &  0:830, 1:851, 2:935,  3:519                       \\
		& $|\EM^\tar|$ & 18982                             & 16552     &      13801                  \\
		\midrule
		\multirow{4}{*}{\bf{DBLP}} & $|\VM^\src|$ & 0:3067, 1:7278, 2:1598, 3:57 &   0:1558, 1:4035, 2:956, 3:47      &    0:1416, 1:3385, 2:820, 3:31     \\
		& $|\EM^\src|$ & 118063                                    &  51415     &      39016                      \\
		& $|\VM^\tar|$ & 0:3067, 1:7278, 2:1598, 3:57 &  0:1654, 1:4083, 2:951, 3:43      &  0:1453, 1:3559, 2:825,  3:48                        \\
		& $|\EM^\tar|$ & 118063                      & 50847     &      45727                  \\
		\bottomrule   
	\end{tabular}
	\caption{Statistics of heterogeneous graphs used in our experiments. The information of node types is included. For example, 0:348 means 348 nodes in this graph are of type 0. 
	}\label{tab:hete datasets}
\end{table*}

\subsection{Complexity Analysis}\label{sec:complexity}
The computational costs can be divided into four parts.
{\bf (i)~}The cost for calculating the $K$\textsuperscript{th}-order proximity matrices $\CB^\src$ and $\CB^\tar$ is $\OM(KVE)$ where $V=\max\{|\VM^\src|,|\VM^\tar|\}$ and $E=\max\{|\EM^\src|,|\EM^\tar|\}$.
{\bf (ii)~}When updating the transport plan, the gradient of $f^{r}\big(\TB^{r},\ZB^{\src,(m)},\ZB^{\tar,(m)}\big)$ in (\ref{eq:pot}) takes the form \cite{peyre2016gromov,xu2019gromov}
\begin{equation}\label{eq:GW gradient}
	\begin{aligned}
		\nabla f^{r}(\TB^{r,(m)},\ZB^{\src,(m)},&\ZB^{\tar,(m)})=h(\CB^\src)\YB^{r,(m)}\oneB^{|\VM^\tar|}\oneB^{|\VM^\tar|\top}\\
		+&\oneB^{|\VM^\src|}\Big(\YB^{r,(m)\top}\oneB^{|\VM^\src|}\Big)^{\top} h(\CB^\tar)\\
		-&2\CB^\src\YB^{r,(m)} \CB^\tar-\alpha\KB^{(m)},
	\end{aligned}
\end{equation}
where $h(\cdot)$ is the element-wise square operation and we use matrix notations $\YB^{r,(m)}=\TB^{r,(m)}+2\sum_{a=1}^{r-1}\hat{\TB}^{a}$ and $\KB^{(m)}=[\kappa(\zB_{i}^{\src,(m)},\zB_{i'}^{\tar,(m)})]$.
Adopting the $k$-rank approximations for $\CB^\src$ and $\CB^\tar$, the cost for computing $\nabla f^{r}(\TB^{r,(m)},\ZB^{\src,(m)},\ZB^{\tar,(m)})$ is $\OM\big((k+d)V^2\big)$ \cite{scetbon2021linear}, where $d$ is the dimension of node embeddings.
The complexity for obtaining the $k$-rank approximation for $\CB^\src$ and $\CB^\tar$ via SVD is $\OM(kV^2)$ \cite{golub1989matrix,halko2011finding}.
Problem (\ref{eq:pot}) can be solved by \emph{mirror descent} \cite{bubeck2015convex,peyre2016gromov} which involves \emph{iterative Bregman projections} \cite{benamou2015iterative}.
If we run mirror descent for $N$ iterations in total for learning the transport plan, each of which involves $T$ matrix-vector multiplications in the projection, the complexity for updating the transport plan is $\OM\big(N(T+k+d)V^2\big)$.
{\bf (iii)~}For learning the embeddings, by selecting the size of node batch as $B\ll V$, the
complexity for updating the embeddings is $\OM(VBd)$ \cite{xu2019gromov} and can be ignored compared to that of learning the transport plan.
{\bf (iv)~}With reasonable $\delta$ and $n$, the complexity for line search is of the same order as for learning the transport plan.
Therefore, the overall complexity is $\OM\big(KVE+N(T+k+d)V^2\big)$.

\begin{table*}[ht]
	\small
	\centering
	\begin{tabular}{cl|ccc|ccc|ccc}
		\toprule
		\multirow{2}{*}{$\rho$} & \multirow{2}{*}{\textbf{Methods}} & \multicolumn{3}{c|}{\textbf{Arenas Email}} & \multicolumn{3}{c|}{\textbf{PPI Yeast}} &  \multicolumn{3}{c}{\textbf{Arxiv}} \\
		&                          & recall      & precision      & F1      & reccall     & precision     & F1     & recall    & precision    & F1   \\
		\midrule
		\multirow{9}{*}{1.0} & \textbf{REGAL} & 97.3$\pm$0.0      & 97.3$\pm$0.0      & 97.3$\pm$0.0      & 81.1$\pm$0.0 		   & 81.1$\pm$0.0 			& 81.1$\pm$0.0       	& 77.9$\pm$0.0 			& 77.9$\pm$0.0 				& 77.9$\pm$0.0\\
		& \textbf{GDD}                                       & 24.2$\pm$0.0      & 24.2$\pm$0.0      & 24.2$\pm$0.0     & 28.3$\pm$0.0 			& 28.3$\pm$0.0 		& 28.3$\pm$0.0     	& 18.7$\pm$0.0			& 18.7$\pm$0.0				& 18.7$\pm$0.0\\
		& \textbf{GRAMPA}                              & 40.9$\pm$0.0     & 40.9$\pm$0.0      & 40.9$\pm$0.0       	  & 32.4$\pm$0.0 		& 32.4$\pm$0.0 		& 32.4$\pm$0.0       &  /                              &    /                          &   /             						  \\
		& \textbf{GWL}                                     & 95.4$\pm$0.3       & 95.4$\pm$0.3       & 95.4$\pm$0.3        & 84.9$\pm$0.6 		& 84.9$\pm$0.6 	  & 84.9$\pm$0.6      	& 78.5$\pm$0.1 				& 78.5$\pm$0.1 			& 78.5$\pm$0.1\\
		& \textbf{MM}                                       & 97.3$\pm$0.0       & 97.3$\pm$0.0       & 97.3$\pm$0.0       	& 80.9$\pm$0.0 		  & 80.9$\pm$0.0 	& 80.9$\pm$0.0       	& 77.8$\pm$0.0 			& 77.8$\pm$0.0 			& 77.8$\pm$0.0\\
		& \textbf{SpectralPivot}                        & 97.4$\pm$0.0        & 97.4$\pm$0.0        & 97.4$\pm$0.0        & 84.0$\pm$0.0        & 84.0$\pm$0.0 	& 84.0$\pm$0.0      	& 73.5$\pm$0.0 			& 73.5$\pm$0.0 			& 73.5$\pm$0.0\\
		& \textbf{\name{}-GW}            				& 97.9$\pm$0.0        & 97.9$\pm$0.0        & 97.9$\pm$0.0      & 85.5$\pm$0.0     & 85.5$\pm$0.0        & 85.5$\pm$0.0     	& 79.0$\pm$0.0 			& 79.0$\pm$0.0 			& 79.0$\pm$0.0\\
		& \textbf{\name{}-W} 							& 5.1$\pm$0.3 			& 5.1$\pm$0.3 			& 5.1$\pm$0.3 		& 2.6$\pm$0.2 			& 2.6$\pm$0.2 			& 2.6$\pm$0.2			& 0.5$\pm$0.0				& 0.5$\pm$0.0			& 0.5$\pm$0.0\\
		& \textbf{\name{}}                                 & \bf{98.4$\pm$0.0} & \bf{98.4$\pm$0.0} & \bf{98.4$\pm$0.0} & \bf{86.5$\pm$0.1} & \bf{86.5$\pm$0.1} & \bf{86.5$\pm$0.1} & \bf{79.1$\pm$0.0} &\bf{ 79.1$\pm$0.0} & \bf{79.1$\pm$0.0}\\
		\midrule
		\multirow{9}{*}{0.8} & \textbf{REGAL}   & 28.4$\pm$0.7        & 25.1$\pm$0.6       & 26.6$\pm$0.7		  & 27.4$\pm$1.1 		& 24.5$\pm$1.0 		& 25.9$\pm$1.0 	& 24.1$\pm$0.1 			& 21.7$\pm$0.1 			& 22.8$\pm$0.1\\
		& \textbf{GDD}                                      & 2.4$\pm$0.0          & 2.1$\pm$0.0        & 2.3$\pm$0.0			& 3.5$\pm$0.0 		  & 3.1$\pm$0.0 		& 3.3$\pm$0.0			& 0.4$\pm$0.0 			& 0.3$\pm$0.0 			& 0.3$\pm$0.0\\
		& \textbf{GRAMPA}                              & 9.3$\pm$0.0           & 8.2$\pm$0.0        & 8.7$\pm$0.0  			& 15.5$\pm$0.0 		  & 13.8$\pm$0.0 	  & 14.6$\pm$0.0		&     /         					 &     /         					&   /             \\
		& \textbf{GWL}                                     & \bf{91.8$\pm$0.8}  & 81.5$\pm$1.4       & 86.4$\pm$0.9 		  & 55.3$\pm$1.3 		& 49.3$\pm$1.2 		& 52.1$\pm$1.2		& 62.7$\pm$0.2 			& 56.2$\pm$0.2 		& 59.3$\pm$0.2\\
		& \textbf{MM}                                       & 26.0$\pm$0.9        & 22.9$\pm$0.8      & 24.3$\pm$0.9		 & 25.1$\pm$1.3 		& 22.3$\pm$1.2 		& 23.6$\pm$1.3			& 22.0$\pm$0.1 			& 19.7$\pm$0.1 			& 20.8$\pm$0.1\\
		& \textbf{SpectralPivot}                    & 89.3$\pm$0.1 		& 78.7$\pm$0.1       & 83.6$\pm$0.1		  & 67.6$\pm$0.8 		& 60.2$\pm$0.7		 & 63.7$\pm$0.7			& 16.4$\pm$0.3 & 14.7$\pm$0.3		& 15.5$\pm$0.3\\
		& \textbf{\name{}-GW}                 			& 89.8$\pm$0.0        & 95.2$\pm$0.0      & 92.4$\pm$0.0 		 & 68.8$\pm$0.0	  & 67.6$\pm$0.0 		& 68.2$\pm$0.0				& 70.4$\pm$0.0 & \bf{79.2$\pm$0.0} & 74.6$\pm$0.0\\
		& \textbf{\name{}-W} 							& 0.8$\pm$0.0 			& 0.8$\pm$0.0 		& 0.8$\pm$0.0			& 0.2$\pm$0.0 		& 0.2$\pm$0.0 			&0.2$\pm$0.0 & 0.2$\pm$0.1 & 0.2$\pm$0.1 & 0.2$\pm$0.1\\
		& \textbf{\name{}}                                 & 90.6$\pm$0.1        & \bf{96.8$\pm$0.7} & \bf{93.6$\pm$0.3} & \bf{69.5$\pm$0.0} & \bf{67.7$\pm$0.1} & \bf{68.6$\pm$0.0} & \bf{71.2$\pm$0.1} & 79.2$\pm$0.1 & \bf{75.0$\pm$0.1}\\
		\midrule
		\multirow{9}{*}{0.6} & \textbf{REGAL}    & 7.0$\pm$0.5       & 5.3$\pm$0.4          & 6.0$\pm$0.5			& 19.1$\pm$0.9 		& 14.7$\pm$0.7 		  & 16.6$\pm$0.8		& 9.7$\pm$0.1 			& 7.4$\pm$0.1 			& 8.4$\pm$0.1\\
		& \textbf{GDD}                                      & 1.2$\pm$0.0         & 0.9$\pm$0.0         & 1.1$\pm$0.0 			& 1.9$\pm$0.0 		  & 1.4$\pm$0.0 		& 1.6$\pm$0.0				& 0.2$\pm$0.0 			& 0.2$\pm$0.0 			& 0.2$\pm$0.0\\
		& \textbf{GRAMPA}                               & 1.1$\pm$0.0          & 0.8$\pm$0.0         & 0.9$\pm$0.0  		&5.8$\pm$0.0 		 & 4.5$\pm$0.0 		& 5.0$\pm$0.0 				&     /          				&     /         					&   /           				  \\
		& \textbf{GWL}                                      & 4.5$\pm$2.7        & 3.5$\pm$1.9          & 3.5$\pm$1.1 			& 34.9$\pm$0.6 		& 26.8$\pm$0.4 		& 30.3$\pm$0.5			& 53.2$\pm$0.6 		& 40.6$\pm$0.5 			& 46.1$\pm$0.5\\
		& \textbf{MM}                                        & 4.6$\pm$0.5        & 3.5$\pm$0.4         & 3.9$\pm$0.4		  & 17.3$\pm$1.0 		 & 13.3$\pm$0.8 		& 15.1$\pm$0.9			& 7.8$\pm$0.1 			& 6.0$\pm$0.1 			& 6.8$\pm$0.1\\
		& \textbf{SpectralPivot}                         & 10.7$\pm$1.6        & 8.1$\pm$1.2          & 9.2$\pm$1.4			& 33.6$\pm$0.7	 	 & 25.8$\pm$0.5 	    & 29.2$\pm$0.6 		& 6.2$\pm$0.2 & 4.7$\pm$0.2 & 5.4$\pm$0.2\\
		& \textbf{\name{}-GW}                  			& 15.0$\pm$0.0      & 15.2$\pm$0.0      & 15.1$\pm$0.0 		   & 42.1$\pm$0.0 		& 41.3$\pm$0.0        & 41.7$\pm$0.0 		& 62.6$\pm$0.0 & 68.6$\pm$0.0 & 65.5$\pm$0.0\\
		& \textbf{\name{}-W} 							& 0.0$\pm$0.0 		& 0.0$\pm$0.0 			& 0.0$\pm$0.0 		 & 0.4$\pm$0.0			& 0.4$\pm$0.0		& 0.4$\pm$0.0 & 0.2$\pm$0.0 & 0.2$\pm$0.0 & 0.2$\pm$0.0\\
		& \textbf{\name{}}                                 & \bf{17.3$\pm$0.3} & \bf{17.4$\pm$0.3} & \bf{17.3$\pm$0.3} & \bf{53.0$\pm$0.2} &\bf{50.2$\pm$0.2} & \bf{51.6$\pm$0.2} & \bf{63.2$\pm$0.2} & \bf{70.2$\pm$0.0} & \bf{66.5$\pm$0.1}\\
		\bottomrule
	\end{tabular}
	\caption{Recall, precision and \Fone{} scores with standard deviations on homogeneous graphs (in percent). The similarity matrix in GRAMPA incurs quadruple computational complexity and takes thousands of hours to obtain on Arxiv.
		Its performance is thus not reported on this \dataset{}.}\label{tab:homo results}
\end{table*}

%% file: sections/experiment.tex
We compare \name{} with state-of-the-art methods on both homogeneous and heterogeneous graphs.
The experiments are conducted on a Ubuntu 18.04 server with a 24-core 2.70GHz Intel Xeon Platinum 8163 CPU, an NVIDIA Tesla V100 GPU, and 92 GB RAM.
The source code is written in Python 3.6 and C++.

\begin{table*}[ht]
	\centering
	\small
		\begin{tabular}{cl|ccc|ccc|ccc}
			\toprule
			\multirow{2}{*}{$\rho$} & \multirow{2}{*}{\bf{Methods}} & \multicolumn{3}{|c}{\bf{Movie}}  & \multicolumn{3}{|c}{\bf{PubMed}} & \multicolumn{3}{|c}{\bf{DBLP}} \\
			&                          & recall  & precision  & F1   & recall  & precision & F1 & recall  & precision & F1  \\
			\midrule
			\multirow{12}{*}{1.0} & \bf{REGAL}   & 73.7$\pm$0.0 		& 73.7$\pm$0.0 		& 73.7$\pm$0.0 		& 60.3$\pm$0.0 		& 60.3$\pm$0.0 			& 60.3$\pm$0.0 			& 81.0$\pm$0.0 			& 81.0$\pm$0.0 		& 81.0$\pm$0.0 				\\
			& \bf{GDD}                      					& 21.9$\pm$0.0 			& 21.9$\pm$0.0 		& 21.9$\pm$0.0  	& 16.4$\pm$0.0 			& 16.4$\pm$0.0    	&61.3$\pm$0.0	& 17.7$\pm$0.0 			& 17.7$\pm$0.0 			& 17.7$\pm$0.0				\\
			& \bf{GRAMPA}                   				& 12.3$\pm$0.0 			& 12.3$\pm$0.0 		& 12.3$\pm$0.0  	& / 								& / 							& / 							&/ 									&/ 									& /									\\
			& \bf{GWL}                      					& 88.2$\pm$1.4 			& 88.2$\pm$1.4 		& 88.2$\pm$1.4    	& 61.2$\pm$0.8 		& 61.2$\pm$0.8 			& 61.2$\pm$0.8  	& 81.6$\pm$0.1 			& 81.6$\pm$0.1 			& 81.6$\pm$0.1			\\
			& \bf{MM}                       					& 73.5$\pm$0.1 			& 73.5$\pm$0.1 		& 73.5$\pm$0.1   	& 60.2$\pm$0.0 		& 60.2$\pm$0.0 			& 60.2$\pm$0.0 		& 81.0$\pm$0.0 			& 81.0$\pm$0.0 			& 81.0$\pm$0.0				\\
			& \bf{SpectralPivot} 							& 87.6$\pm$0.5		& 87.6$\pm$0.5		 & 87.6$\pm$0.5 		& 59.6$\pm$0.7		&59.6$\pm$0.7 			& 59.6$\pm$0.7			&78.9$\pm$0.1	&78.9$\pm$0.1			&78.9$\pm$0.1			\\
			& \bf{SANA}                							& 65.8$\pm$0.5		& 65.8$\pm$0.5		& 65.8$\pm$0.5			&/									&/									&/									&/								&/									&/									\\
			& \bf{VELSET}                                    & 4.7$\pm$0.0          & 4.7$\pm$0.0         & 4.7$\pm$0.0          & 5.7$\pm$0.0          &5.7$\pm$0.0             &5.7$\pm$0.0      &0.5$\pm$0.0             & 0.5$\pm$0.0          & 0.5$\pm$0.0\\
			& \bf{G-Finder}                                  & 1.7$\pm$0.0           & 3.3$\pm$0.0         & 2.2$\pm$0.0          &/                            &/                                &/                         &/                               &/                               &/                   \\
			& \bf{\name{}-GW}                   		& 91.6$\pm$0.0 			& 91.6$\pm$0.0 		& 91.6$\pm$0.0 			& 65.5$\pm$0.0	&65.5$\pm$0.0			&65.5$\pm$0.0			& 81.2$\pm$0.0		&81.2$\pm$0.0				&81.2$\pm$0.0 		\\
			& \bf{\name{}-W}							& 17.3$\pm$3.6			&17.3$\pm$3.6			&17.3$\pm$3.6 			&1.1$\pm$0.1				&1.1$\pm$0.1			&1.1$\pm$0.1			& 0.3$\pm$0.0			& 0.3$\pm$0.0			& 0.3$\pm$0.0		\\
			& \bf{\name{}}                     				& \bf{93.7$\pm$0.3} & \bf{93.7$\pm$0.3} & \bf{93.7$\pm$0.3}	& \bf{68.0$\pm$0.0} &\bf{68.0$\pm$0.0}&\bf{68.0$\pm$0.0} & \bf{81.6$\pm$0.0} & \bf{81.6$\pm$0.0} & \bf{81.6$\pm$0.0} 	  \\
			\midrule
			\multirow{12}{*}{0.8} & \bf{REGAL}   & 32.8$\pm$1.2 			& 29.6$\pm$1.1 			& 31.1$\pm$1.1 	& 35.0$\pm$0.3		& 31.5$\pm$0.3 		& 33.2$\pm$0.3 			& 19.5$\pm$0.3 			& 17.5$\pm$0.3 	& 18.5$\pm$0.3			\\
			& \bf{GDD}                      					& 2.9$\pm$0.0 			& 2.6$\pm$0.0 			& 2.7$\pm$0.0 		& 1.7$\pm$0.0 			& 1.5$\pm$0.0 		& 1.6$\pm$0.0   		& 0.6$\pm$0.0 			& 0.5$\pm$0.0 		& 0.5$\pm$0.0			\\
			& \bf{GRAMPA}                   				& 6.0$\pm$0.0 			& 5.4$\pm$0.0 			& 5.7$\pm$0.0 		&/								&/								&/									&/										&/							&/									\\
			& \bf{GWL}                      					& 73.0$\pm$2.6 			& 66.0$\pm$2.3 		& 69.3$\pm$2.4 		& 48.1$\pm$0.2 			& 43.3$\pm$0.1		&45.6$\pm$0.1  		& 39.3$\pm$1.5 			& 34.9$\pm$1.1 		& 37.0$\pm$1.3			\\
			& \bf{MM}                       					& 29.3$\pm$1.4 		& 26.5$\pm$1.3 				& 27.8$\pm$1.3   & 33.5$\pm$0.6 			& 30.1$\pm$0.6 			&31.7$\pm$0.6 		& 16.0$\pm$0.6 			& 14.3$\pm$0.5 		& 15.1$\pm$0.5 		\\
			& \bf{SpectralPivot}                   		& 72.0$\pm$1.9			& 65.0$\pm$1.7			& 68.3$\pm$1.8 		&47.7$\pm$0.4			&43.0$\pm$0.4			&45.2$\pm$0.4		&24.5$\pm$9.4			&21.9$\pm$8.4		&23.1$\pm$8.8		\\
			& \bf{SANA}                						& 49.8$\pm$1.5 			& 45.0$\pm$1.4 			& 47.3$\pm$1.5   	& /					 			&/								 	&/									& /									& / 							& / 							\\
			& \bf{VELSET}                                 & 4.4$\pm$0.0             & 4.0$\pm$0.0            & 4.1$\pm$0.0         &4.5$\pm$0.0        &4.1$\pm$0.0            &4.3$\pm$0.0              & 0.5$\pm$0.0            &0.4$\pm$0.0                 &0.4$\pm$0.0\\
			& \bf{G-Finder}                                & 1.1$\pm$0.0             & 3.1$\pm$0.0            & 1.6$\pm$0.0          &/                         &/                               &/                                 &/                                &/                                    &/ \\
			& \bf{\name{}-GW}                  			& 74.4$\pm$0.0			& 71.9$\pm$0.0		 & 73.1$\pm$0.0		&49.3$\pm$0.0			&50.5$\pm$0.0		&49.9$\pm$0.0			& 32.9$\pm$0.0                      & 46.6$\pm$0.0		& 38.6$\pm$0.0		\\
			& \bf{\name{}-W}                     		& 2.0$\pm$0.1				& 2.7$\pm$0.4		 & 2.3$\pm$0.2		&0.5$\pm$0.1			&0.5$\pm$0.1			&0.5$\pm$0.1 			& 0.5$\pm$0.0				&0.5$\pm$0.0		&0.5$\pm$0.0				\\
			& \bf{\name{}}									& \bf{75.7$\pm$0.0}	& \bf{72.2$\pm$0.2}	&\bf{73.9$\pm$0.1} &\bf{53.7$\pm$0.0} &\bf{52.6$\pm$0.0} &\bf{53.1$\pm$0.0} & \bf{41.7$\pm$1.0} & \bf{59.7$\pm$1.3} & \bf{49.1$\pm$1.2} \\
			\midrule
			\multirow{12}{*}{0.6} & \bf{REGAL}  & 19.5$\pm$0.7		& 14.8$\pm$0.6 		& 16.9$\pm$0.6 		& 18.1$\pm$0.5 		& 14.0$\pm$0.4 			& 15.8$\pm$0.4  		& 10.4$\pm$0.2 		& 7.9$\pm$0.2 			& 9.0$\pm$0.2 			\\
			& \bf{GDD}                      					& 2.1$\pm$0.0 		& 1.6$\pm$0.0 			& 1.8$\pm$0.0  		& 0.8$\pm$0.0 		& 0.6$\pm$0.0 			& 0.7$\pm$0.0  			& 0.6$\pm$0.0 			& 0.4$\pm$0.0 		& 0.5$\pm$0.0 			\\
			& \bf{GRAMPA}                   				& 3.1$\pm$0.0 		& 2.4$\pm$0.0 			& 2.7$\pm$0.0 		&/							&/									&/									&/									&/								&/								\\
			& \bf{GWL}                      					& 60.6$\pm$3.5		& 46.1$\pm$2.7 		& 52.4$\pm$3.0  	& 35.2$\pm$0.2	& 27.2$\pm$0.1 			& 30.7$\pm$0.2			& \bf{29.3$\pm$0.3} & 22.3$\pm$0.2 		& 25.3$\pm$0.3		\\
			& \bf{MM}                       					& 14.9$\pm$0.7 		& 11.3$\pm$0.5 			& 12.9$\pm$0.6 		& 15.4$\pm$0.3 		& 11.9$\pm$0.3 			& 13.4$\pm$0.3			& 7.8$\pm$0.6 			& 5.9$\pm$0.4 		& 6.8$\pm$0.5			\\
			& \bf{SpectralPivot}                   			& 41.1$\pm$2.6		& 31.2$\pm$2.0		 & 35.5$\pm$2.2		&33.0$\pm$0.8		&25.5$\pm$0.6			&28.7$\pm$0.7			&17.8$\pm$1.4		&13.5$\pm$1.1			&15.4$\pm$1.2		\\
			& \bf{SANA}                							& 30.1$\pm$1.3 		& 22.9$\pm$1.0 		& 26.0$\pm$1.1   	& /								&/									&/									&/							&/									&/									\\
			& \bf{VELSET}                                    & 4.2$\pm$0.0        & 3.2$\pm$0.0         & 3.6$\pm$0.0      &3.8$\pm$0.0            &2.9$\pm$0.0           &3.3$\pm$0.0          &0.4$\pm$0.0            &0.3$\pm$0.0           &0.3$\pm$0.0 \\
			& \bf{G-Finder}                                  & 0.8$\pm$0.0         &2.2$\pm$0.0         &1.2$\pm$0.0        & /                           & /                                & /                            & /                            & /                             & /                    \\
			& \bf{\name{}-GW}                 			& 76.1$\pm$0.0		& 62.9$\pm$0.0		 & 68.9$\pm$0.0  	&33.3$\pm$0.0		&29.9$\pm$0.0			&31.5$\pm$0.0 			& 27.0$\pm$0.0	&31.0$\pm$0.0			& 28.8$\pm$0.0		\\
			& \bf{\name{}-W}                     			&1.1$\pm$0.4		&0.9$\pm$0.3		 &1.0$\pm$0.4 			&0.3$\pm$0.1			&0.3$\pm$0.1			& 0.3$\pm$0.1	  		& 0.1$\pm$0.0			& 0.1$\pm$0.0			& 0.1$\pm$0.0 			\\
			& \bf{\name{}} 									&\bf{77.9$\pm$0.3} &\bf{63.7$\pm$0.1} &\bf{70.1$\pm$0.2} & \bf{37.1$\pm$0.0} &\bf{33.3$\pm$0.0} & \bf{35.1$\pm$0.0} & 27.9$\pm$0.1		& \bf{32.7$\pm$0.2} & \bf{30.1$\pm$0.1} \\
			\bottomrule
		\end{tabular}
	\caption{Recall, precision, and \Fone{} type mismatch ratio $q$ scores with standard deviations on heterogeneous graphs (in percent).}\label{tab:hete results}
\end{table*}

\begin{table*}[ht]
	\small
	\centering
	\begin{tabular}{l|ccc|ccc|ccc}
		\toprule
		{\bf Datasets}                    & \multicolumn{3}{c|}{\bf Movie} & \multicolumn{3}{c|}{\bf PubMed} & \multicolumn{3}{c}{\bf DBLP} \\
		$\rho$                     & 1.0     & 0.8    & 0.6    & 1.0     & 0.8     & 0.6    & 1.0      & 0.8     & 0.6     \\
		\midrule
		{\bf REGAL}            & 12.2$\pm$0.9    & 27.4$\pm$1.5     & 34.2$\pm$1.3     & 29.1$\pm$0.5    & 50.9$\pm$0.4     & 63.0$\pm$1.0    & 0.1$\pm$0.0       &3.4$\pm$0.1        & 6.6$\pm$0.3        \\
		{\bf GDD}               & 37.9$\pm$0.0    & 55.9$\pm$0.0     & 55.9$\pm$0.0    & 61.3$\pm$0.0    & 72.8$\pm$0.0     &  73.4$\pm$0.0   &  33.8$\pm$0.0   & 49.5$\pm$0.0     &  52.1$\pm$0.0       \\
		{\bf GRAMPA}        & 50.5$\pm$0.0    & 57.8$\pm$0.0     & 51.5$\pm$0.0     &   /                      &  /                         &  /                       &  /                        &  /                         &  /       \\
		{\bf GWL}              & 4.5$\pm$0.8      & 16.5$\pm$1.4      & 25.2$\pm$1.6     & 28.6$\pm$0.7    &  42.9$\pm$0.2    & 54.7$\pm$0.2   &  0.1$\pm$0.0      &  7.2$\pm$0.8      &  5.0$\pm$0.6       \\
		{\bf MM}                & 10.9$\pm$0.0    & 37.1$\pm$2.0      & 47.5$\pm$0.9     & 29.1$\pm$0.5    &  51.8$\pm$0.6     & 64.4$\pm$1.0   & 0.3$\pm$0.0       & 19.2$\pm$0.9     & 24.2$\pm$0.9        \\
		{\bf SpectralPivot} & 4.8$\pm$0.3      & 16.7$\pm$1.1       & 32.3$\pm$2.0    & 30.5$\pm$0.4    & 42.9$\pm$0.5     &  55.3$\pm$0.3  &  4.5$\pm$0.3      &  21.0$\pm$6.3    &  20.7$\pm$0.3       \\
		{\bf SANA}            & 10.0$\pm$0.6     & 6.9$\pm$0.7       & 8.9$\pm$0.5       &    /                     &  /                         &  /                       &   /                       &  /                         & /        \\
		{\bf VELSET}         &{\bf 0.0$\pm$0.0}&{\bf 0.0$\pm$0.0}& 2.1$\pm$0.0      &{\bf 0.0$\pm$0.0}&{\bf 0.0$\pm$0.0}& 0.1$\pm$0.0      & 0.1$\pm$0.0       & 0.1$\pm$0.0       &  0.3$\pm$0.0       \\
		{\bf G-Finder}       &{\bf 0.0$\pm$0.0}&{\bf 0.0$\pm$0.0}&{\bf 0.0$\pm$0.0}&/                          &/                          & /                         & /                          & /                         & /        \\
		{\bf \name{}-GW} &{\bf 0.0$\pm$0.0}&{\bf 0.0$\pm$0.0}&{\bf 0.0$\pm$0.0}&{\bf 0.0$\pm$0.0}&{\bf 0.0$\pm$0.0}&{\bf 0.0$\pm$0.0}&{\bf 0.0$\pm$0.0}&{\bf 0.0$\pm$0.0}&{\bf 0.0$\pm$0.0}         \\
		{\bf \name{}-W}   &{\bf 0.0$\pm$0.0}&{\bf 0.0$\pm$0.0}&{\bf 0.0$\pm$0.0}&{\bf 0.0$\pm$0.0}&{\bf 0.0$\pm$0.0}&{\bf 0.0$\pm$0.0}&{\bf 0.0$\pm$0.0}&{\bf 0.0$\pm$0.0}&{\bf 0.0$\pm$0.0}         \\
		{\bf \name{}}        &{\bf 0.0$\pm$0.0}&{\bf 0.0$\pm$0.0}&{\bf 0.0$\pm$0.0}&{\bf 0.0$\pm$0.0}&{\bf 0.0$\pm$0.0}&{\bf 0.0$\pm$0.0}&{\bf 0.0$\pm$0.0}&{\bf 0.0$\pm$0.0}&{\bf 0.0$\pm$0.0}        \\
		\bottomrule
	\end{tabular}
	\caption{Type mismatch ratio $q$ results with standard deviations on heterogeneous graphs (in percent).}\label{tab:type_mismatch}
\end{table*}

\subsection{Experimental Setup}

\paragraph{Baselines.}
The baselines can be divided into three families:
1) State-of-the-art methods for matching homogeneous graphs including REGAL \cite{heimann2018regal}, GDD \cite{scott2021graph}, GRAMPA \cite{fan2020spectral}, GWL \cite{xu2019gromov}, MM \cite{konar2020graph}, SpectralPivot \cite{karakasis2021joint};
2) SANA \cite{gu2018homogeneous}, which extends \citet{mamano2017sana} by adopting colored graphlet degree vector features to match heterogeneous graphs\footnote{\citeauthor{gu2018homogeneous}~(\citeyear{gu2018homogeneous}) extend three homogeneous graph matching methods to the heterogeneous variants. SANA is shown to achieve the best performance among them.};
3) Methods that treat heterogeneous graphs as \emph{homogeneous attributed} graphs, including VELSET \cite{dutta2017neighbor} and G-Finder \cite{liu2019g}.
We also conduct ablation studies.
The variants of \name{} include 1) \name{}-GW which only uses the partial GW discrepancy by setting $\alpha=0$ and thus does not involve the embedding learning, and 2) \name{}-W which only uses the partial Wasserstein discrepancy.

\paragraph{Metrics.}
For evaluation of partial graph matching,
we compute the commonly used indicators (see e.g. \cite{sarlin2020superglue,wang2019user}),
\begin{equation*}
	\begin{aligned}
		\recall&=\frac{\#\{\text{correct matching}\}}{\#\{\text{\groundtruth{} matching}\}},\\
		\precision&=\frac{\#\{\text{correct matching}\}}{\#\{\text{total predicted matching}\}},\\
		\Fone&=\frac{2\cdot\recall\cdot\precision}{\recall+\precision}.
	\end{aligned}
\end{equation*}
On experiments of matching heterogeneous graphs, we also report the ratio of type mismatch
\begin{equation*}
	q=\frac{|\{(i,\hat{i})|\phi^\src(i)\neq\phi^\tar(\hat{i})\}|}{\#\{\text{\groundtruth{} matching}\}},
\end{equation*}
where $\hat{i}$ is the output counterpart of node $i$ predicted by each method.
We run each method for $5$ times and report both the average values and standard deviations.

\paragraph{\Dataset{} Preparation.}
We extract fully or partially overlapped subgraphs from benchmark \datasets{}.
Mathematically, the overlap ratio is defined as $\rho=|\VM^\src\cap\VM^\tar|/|\VM^\src\cup\VM^\tar|$.
As $\rho$ decreases, the matching problem becomes more difficult.
We verify the efficacy of \name{} on graphs extracted from three homogeneous graphs, including Arenas Email\footnote{\url{http://konect.cc/networks/arenas-email/}}, PPI Yeast\footnote{\url{https://www3.nd.edu/~cone/MAGNA++/}}, and Arxiv\footnote{\url{http://snap.stanford.edu/data/ca-AstroPh.html}}.
We then compare the performance of \name{} against baselines on heterogeneous graphs, including Movie \footnote{\url{https://github.com/eXascaleInfolab/JUST/tree/master/Datasets/Movies}}, PubMed \footnote{\url{https://pubmed.ncbi.nlm.nih.gov/}} and DBLP \footnote{\url{https://dblp.uni-trier.de/}}.
Movie contains four node types, including actors, movies, directors and composers.
PubMed is a network of genes, diseases, chemicals, and species.
DBLP is an academic network containing authors, papers, venues and phrases.
Statistics of extracted subgraphs are listed in Table \ref{tab:homo datasets} and Table \ref{tab:hete datasets}.

\paragraph{Parameter choices.}
The transport mass is selected by line search as is stated in Sec. \ref{sec:linesearch}.
In all experiments, the embedding dimension is set as $d=64$.
Setting $1\times 10^{-7}\le\alpha\le1\times10^{-4}$ for \name{} yields improved performance over  \name{}-GW.
The results in Tables \ref{tab:homo results}, \ref{tab:hete results} and \ref{tab:type_mismatch} are obtained with $\alpha=1\times10^{-5}$.
We tested $\beta$ in $\{1,  0.1, 0.01, 1\times 10^{-3}, 1\times 10^{-4}, 1\times 10^{-5}\}$.
$1\times 10^{-3}\le\beta\le 1$ achieves stable performance and thus we set $\beta=0.01$.

\subsection{Matching Homogeneous Graphs}

The recall, precision, and F1 scores on homogeneous graphs are shown in Table \ref{tab:homo results}.
\name{} and \name{}-GW consistently outperform baselines in terms of the \Fone{} indicator and the advantage becomes more significant with the overlap ratio decreasing, which demonstrates the effectiveness of partial optimal transport.
GWL and SpectralPivot are closest competitors.
Because GWL is a full matching method, it occasionally outperforms \name{} in terms of recall by a small margin.
However, it is widely known that \Fone{} is a better measure as it balances precision and recall \cite{van1979information,fawcett2006introduction}.
The partial optimal transport allows \name{} to significantly boost the precision and \Fone{}.
For partially overlapped graphs, \name{} improves the precision and \Fone{} over the best baseline by \emph{at least} 12.5\% and 7.7\% respectively and often much more.
The ablation study indicates that the superior performance of \name{} is mainly attributed to the partial GW discrepancy, since \name{}-W has unsatisfying results.
Node embeddings improve the matching performance of \name{}.

\subsection{Matching Heterogeneous Graphs}

The performance of \name{} and baselines on heterogeneous graphs are reported in Tables \ref{tab:hete results} and \ref{tab:type_mismatch}.
Due to the complexity of GRAMPA, SANA and G-Finder, we evaluate them only in the first test on the smaller Movie \dataset{} but not in the remaining two tests on the larger PubMed and DBLP graphs.
Other methods are evaluated on all three tests.
\name{} and \name{}-GW outperform baselines in terms of precision and \Fone{} indicators on all \datasets{}.

All homogeneous graph matching methods can match nodes of different types.
On Movie and PubMed, they match about half of the nodes in the source graph to nodes of different types in the target graph.
Therefore, they cannot be directly used to match heterogeneous graphs and the type information should be explicitly considered.
SANA, VELSET and G-Finder outperform these methods in terms of the type mismatch ratio.
However, type mismatch still occurs.
By contrast, type mismatch do not happen for \name{} and its variants.


%% file: sections/related.tex
\paragraph{Matching homogeneous graphs.}
The matching costs $k_{ii'}$ and $d_{ii'jj'}$ in (\ref{eq:formulation}) are critical to the matching accuracy.
In some early works, these costs are based on handcrafted features that rely heavily on expert knowledge \cite{mamano2017sana,heimann2018regal}.
More recently, end-to-end deep learning frameworks for graph matching are proposed to automatically learn the node embedding-based assignment costs \cite{zanfir2018deep,wang2019learning,wang2019functional},
which however are supervised and require a large amount of \groundtruth{} node pairs to be available.
OT-based methods propose to exploit geometrical properties of the metric space of the graph in order to estimate the node correspondence in an unsupervised/semi-supervised manner and thus reduce the demand of data labeling \cite{maretic2019got,xu2019gromov,titouan2019optimal}.

\paragraph{Matching heterogeneous Graphs.}
Many existing methods for matching heterogeneous graphs are supervised \cite{wu2019relation,ren2020scalable,sun2020knowledge,wang2020knowledge}.
Although \citeauthor{zhang2021unsupervised}~\citeyear{zhang2021unsupervised} propose an unsupervised method, the purpose is different from ours.
Concretely, the nodes represent persons captured by different cameras, and are divided into two types according to whether the appearance is clear.
The node type mismatch is allowed or even encouraged, since a person is clear in one camera may not be clear in another.
Some methods \cite{dutta2017neighbor,liu2019g} treat the node type as an one-dimensional node attribute.
Matching heterogeneous graphs is then converted into the problem of matching homogeneous \emph{attributed} graphs, which may still match nodes of different types.
Besides, these methods are often time-consuming due to the complex matching procedures.